%% file: main.tex
\documentclass[10pt,twocolumn,letterpaper]{article}
\usepackage[table]{xcolor}
\usepackage{appendix}
\usepackage{listings}
\usepackage{graphicx}
\usepackage{booktabs} 
\usepackage{caption} 
\usepackage{geometry}
\usepackage{array}
\usepackage{amsmath}
\usepackage{subfigure}
\usepackage{longtable}
\usepackage{abstract}
\usepackage{setspace}
\usepackage{multirow}
\usepackage{diagbox}
\usepackage{enumerate}
\usepackage{float}
\usepackage{gensymb}
\usepackage{microtype}
\usepackage{makecell}
\usepackage{lipsum}
\usepackage{threeparttable}
\usepackage{pythonhighlight}
\usepackage[ruled,vlined]{algorithm2e}
\usepackage{amsfonts}
\usepackage{subfigure}
\usepackage{svg}
\usepackage{textcomp}

\usepackage[pagenumbers]{iccv} % To force page numbers, e.g. for an arXiv version


\definecolor{iccvblue}{rgb}{0.21,0.49,0.74}
\usepackage[pagebackref,breaklinks,colorlinks,allcolors=iccvblue]{hyperref}

\def\paperID{1948} % *** Enter the Paper ID here
\def\confName{ICCV}
\def\confYear{2025}

\title{Constructing Ophthalmic MLLM for Positioning-diagnosis Collaboration Through Clinical Cognitive Chain Reasoning}

\author{Xinyao Liu$^{2,1}$, Diping Song$^{\dagger1}$ \\
$^1$Shanghai Artificial Intelligence Laboratory\quad 
$^2$University of Science and Technology of China \\
\tt\small{
liuxinyao@mail.ustc.edu.cn, songdiping@pjlab.org.cn}
}

\begin{document}
\maketitle
\input{sec/0_abstract}    
\input{sec/1_intro}

\input{sec/2_related_works}
\input{sec/3_fundus_engine}
\input{sec/4_fundusexpert}

\input{sec/5_experiment}

\input{sec/6_conlcusion}
\input{sec/7_acknowledgment}

\input{sec/supplementary}

{
    % \clearpage
    \small
    \bibliographystyle{ieeenat_fullname}
    \bibliography{egbib}
}
\end{document}

%% file: sec/0_abstract.tex
\begin{abstract}

Multimodal large language models (MLLMs) demonstrate significant potential in the field of medical diagnosis. However, they face critical challenges in specialized domains such as ophthalmology, particularly the fragmentation of annotation granularity and inconsistencies in clinical reasoning logic, which hinder precise cross-modal understanding. This paper introduces \textbf{FundusExpert}, an ophthalmology-specific MLLM with integrated positioning-diagnosis reasoning capabilities, along with \textbf{FundusGen}, a dataset constructed through the intelligent \textbf{Fundus-Engine} system.
Fundus-Engine automates localization and leverages MLLM-based semantic expansion to integrate global disease classification, local object detection, and fine-grained feature analysis within a single fundus image. Additionally, by constructing a clinically aligned cognitive chain, it guides the model to generate interpretable reasoning paths.
FundusExpert, fine-tuned with instruction data from FundusGen, achieves the best performance in ophthalmic question-answering tasks, surpassing the average accuracy of the 40B MedRegA by 26.6\%. It also excels in zero-shot report generation tasks, achieving a clinical consistency of 77.0\%, significantly outperforming GPT-4o's 47.6\%. Furthermore, we reveal a scaling law between data quality and model capability ($L \propto N^{0.068}$), demonstrating that the cognitive alignment annotations in FundusGen enhance data utilization efficiency. By integrating region-level localization with diagnostic reasoning chains, our work develops a scalable, clinically-aligned MLLM and explores a pathway toward bridging the visual-language gap in specific MLLMs. Our project can be found at \url{https://github.com/MeteorElf/FundusExpert}.

{
  \renewcommand{\thefootnote}%
    {\fnsymbol{footnote}}
  \footnotetext[0]{$\dagger$Corresponding author.} 
  }
  
\end{abstract}

%% file: sec/1_intro.tex
\section{Introduction}
\label{sec:intro}

\begin{figure*}[ht]
\centering
\includegraphics[width=\textwidth]{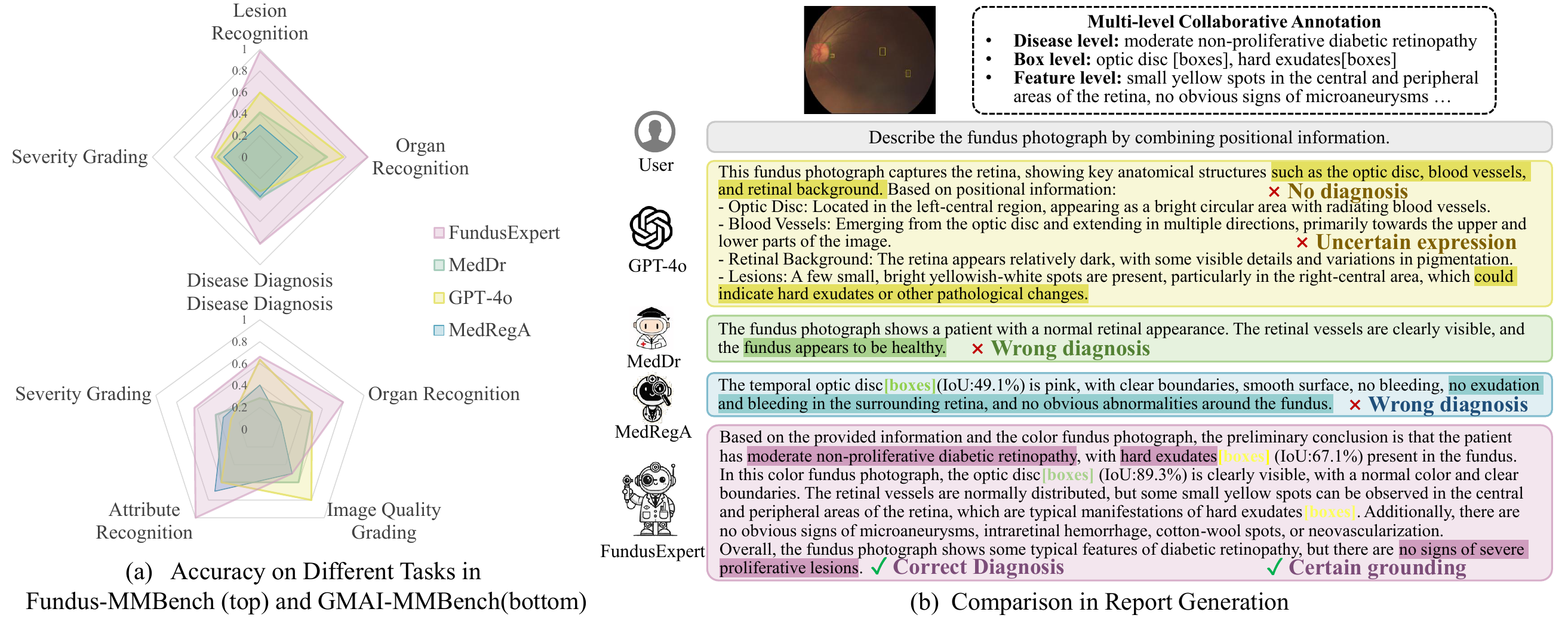}
\caption{Comparison of FundusExpert with other models.}
\label{fig:radar}
\vspace{-2mm}
\end{figure*}

In recent years, multimodal large language models (MLLMs) \cite{chatgpt, bai2023qwen, liu2024visual, chen2024internvl, zhu2023minigpt}demonstrate remarkable performance and generalization potential in cross-modal understanding and reasoning, making them promising tools for computer-aided medical diagnosis.

As a uniquely valuable type of medical imaging, fundus images contain rich lesion features that play a crucial role in ophthalmic disease diagnosis\cite{li2021applications}. Existing general-purpose models \cite{bolton2024biomedlm, he2024meddr, tu2024towards} explore tasks such as medical image-based report generation and disease diagnosis, demonstrating the advantages of MLLMs in cross-modal data processing. However, as illustrated in Fig.~\ref{fig:radar}, these general models often exhibit inferior precision and specificity compared to domain-specific models when applied to specialized medical fields due to the need to handle multiple data modalities simultaneously.  

Although current mainstream multimodal ophthalmology models achieve marginal performance gains through large-scale training with millions of images \cite{shi2024eyeclip, qiu2024development}, their reliance on discretized annotation-based supervision creates misalignment with clinical reasoning processes.
These models are not guided to establish a cognitive chain of \textit{region localization} $\rightarrow$ \textit{feature analysis} $\rightarrow$ \textit{diagnostic reasoning} during training. As shown in Fig.~\ref{fig:radar}, such representational deficiencies lead to spatial perception inaccuracies (e.g., optic disc localization errors in MedRegA \cite{wang2024interpretable}) and disrupted diagnostic correlations (e.g., wrong diagnosis in MedDr \cite{he2024meddr} ).  

The root cause of these deficiencies in spatial perception and diagnostic correlation lies in the significant granularity fragmentation within the multimodal learning framework for ophthalmology. Current mainstream fundus datasets contain global diagnostic labels (e.g., diabetic retinopathy grading) and scattered fine-grained annotations (e.g., local structure segmentation masks), but these annotations often exist in isolation across different dataset subsets. This "single-image-single-dimension" flat mapping paradigm imposes two critical limitations on the model’s ability to analyze complex fundus manifestations. First, the model struggles to establish cross-scale semantic associations, such as mapping microaneurysm distribution at the microscopic level to diabetic retinopathy staging at the macroscopic level. Second, fragmented training objectives fail to simulate the progressive diagnostic pathway adopted by clinicians, which involves analyzing the spatial distribution of fundus structures and local features to infer disease progression.

To mitigate the fragmentation across different annotation granularities and better align with clinical workflows, we construct the FundusGen dataset using the intelligent data system Fundus-Engine. This dataset incorporates a cognitive chain that progresses from local-to-global diagnostic reasoning to in-depth validation of the evidential chain. Fundus-Engine enhances annotation efficiency through automated annotation mechanism, achieving a three-stage collaborative annotation process on a single image. At the model training level, we adopt a task-oriented approach that prioritizes data quality over sheer data scale. We investigate a selective strategy for instruction-tuning data, systematically analyzing how data selection patterns and scale influence model performance to better adapt to ophthalmic tasks.  

Based on the above methodology, we successfully develop a specialized ophthalmic MLLM, FundusExpert. We conduct a comprehensive evaluation of FundusExpert across multiple dimensions of medical-related tasks, including region recognition and detection, clinical question answering, and medical report generation. In clinical question-answering evaluations, FundusExpert outperforms other domain-specific models and proprietary commercial models in both in-distribution and out-of-distribution benchmarks. Specifically, in the GMAI-MMBench\cite{chen2024gmai}, FundusExpert(8B) surpasses the accuracy of the 40B MedRegA by 26.6\%. 
% Furthermore, FundusExpert demonstrates superior performance in ophthalmic medical report generation under zero-shot scenarios and exhibits a stronger grasp of domain knowledge compared to GPT-4o.  This finding inspires an efficient data iteration strategy where synthetic data generation can transition from proprietary commercial models (e.g., GPT-4o) to lightweight specialized models. Synthetic data not only mitigates patient privacy concerns but also significantly reduces annotation costs.  

This paper presents three key contributions:

\begin{enumerate}
    \item \textbf{Proposal of the Fundus-Engine and the construction of a collaborative annotation framework with a clinical cognitive reasoning chain, resulting in the FundusGen dataset.}  
    We achieve collaborative annotation of disease classification, regional localization, and lesion characteristics within a single fundus image, enhancing the association of ophthalmic data across different granularity levels. By aligning with clinical reasoning, we establish a semantic-associative cognitive chain in FundusGen, which expands the model’s disease analysis capabilities and improves interpretability.

    \item \textbf{Development of an ophthalmic MLLM with integrated positioning-diagnosis reasoning: FundusExpert.}  
    This MLLM demonstrates the ability of region-semantic self-reference without the need for external tools. It can associate the spatial location of regions in medical images with corresponding statements in the conversation. Additionally, it implements multi-task progressive inference on fundus color images, offering promising potential as a foundational model for the ophthalmology field.

    \item \textbf{Revealing the scaling law properties of medical multimodal data.}  
    Through subset sampling experiments on FundusGen, we observe a significant scaling law between model performance and data volume. This finding not only quantitatively supports the data quality of FundusGen but also reveals that a granularity-fusion annotation aligned with clinical cognitive logic enhances data utilization efficiency. This provides a reference for constructing specific MLLMs.
\end{enumerate}

%% file: sec/2_related_works.tex
\section{Related Works}

\begin{figure*}[th]
\centering
\includegraphics[width=\textwidth]{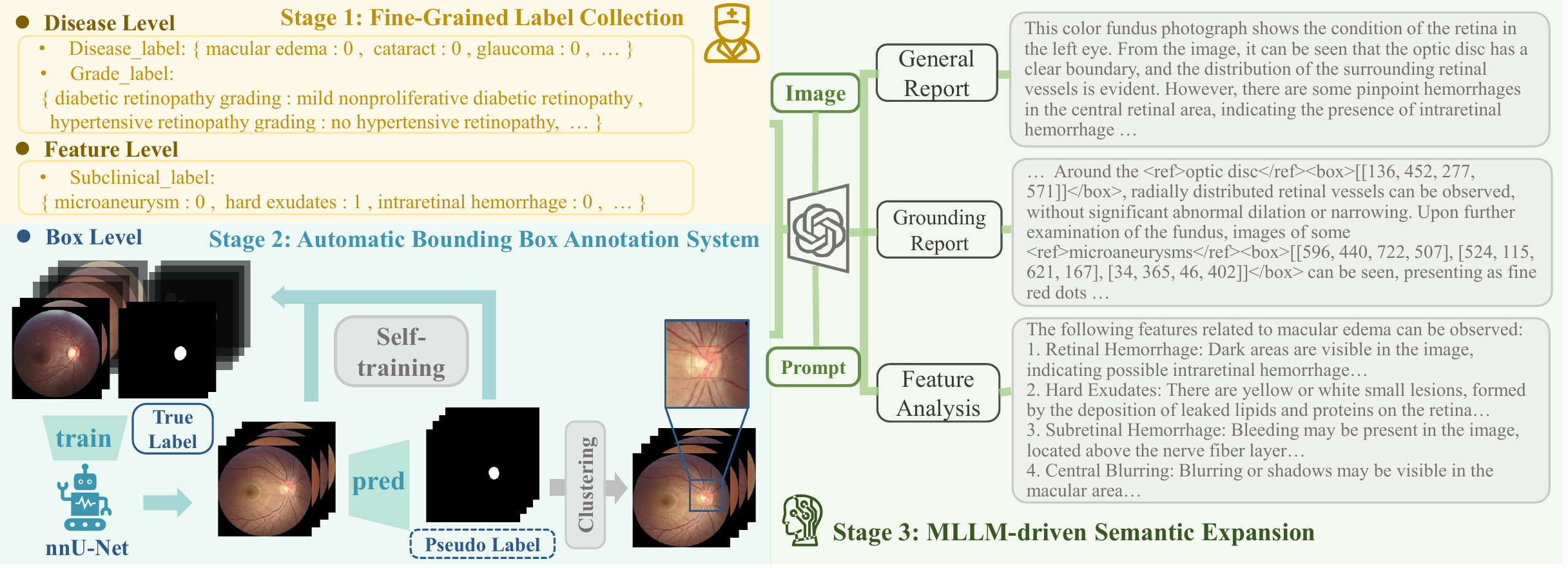}
\caption{Collaborative Annotation of Fundus-Engine.}
\label{fig:data_pipeline}
\vspace{-2mm}
\end{figure*}

\noindent\textbf{Ophthalmic Supervised Learning Paradigms.}
VisionFM \cite{qiu2024development} constructs a multimodal foundation model based on 3.4 million ophthalmic images, achieving marginal improvements in performance. RetiZero \cite{wang2024common} employs contrastive pretraining on 340,000 eye fundus image-text pairs, covering over 400 diseases, but its global alignment strategy is susceptible to interference from false negative samples (e.g., images and texts that are semantically identical but misclassified as negative pairs). To address this issue, ViLReF \cite{yang2024vilref} proposes the Weighted Similarity Coupling Loss and dynamic memory queue, which guide label extraction and compensate for the absence of false negative samples using expert knowledge. However, such methods still rely on a “single image-single label” coarse-grained alignment, lacking hierarchical semantic connections in the annotation system and depending on large-scale supervised learning paradigms.

\noindent\textbf{Ophthalmic-Specific Multimodal Models.}
RETFound \cite{zhou2023foundation} extracts general representations from 1.6 million unlabeled eye fundus images through self-supervised learning, but its pretraining process does not integrate the text modality, making it difficult for the model to establish fine-grained associations between visual features and clinical descriptions. DeepDR-LLM \cite{li2024integrated} combines 372,000 primary care chronic disease diagnosis and management data to optimize LLM training. However, the model's training tasks primarily focus on diabetic retinopathy grading, making it challenging to extend to other fundus diseases. VisionUnite \cite{li2024visionunite} fine-tunes with 296,379 high-quality eye fundus image-text pairs. However, its text data is largely generated by GPT-4V, leading to limitations in data quality and the accuracy of medical knowledge.

\noindent\textbf{Region-Aware Medical MLLMs.}
Enhancing the model's region-aware capabilities for medical images has become a critical research direction\cite{xie2024medtrinity}. General medical MLLMs such as MedDr \cite{he2024meddr} improve disease diagnosis accuracy through expert collaboration mechanisms, but their region-awareness relies on external tools, making semantic self-reference challenging. In the ophthalmic domain, VisionFM \cite{qiu2024development} adopts a two-stage "detection followed by description" paradigm, which results in a disconnection between region analysis and diagnostic decisions. This decoupled design deviates from the integrated cognitive logic of clinical doctors, limiting the model's reliability and interpretability in complex scenarios.

%% file: sec/3_fundus_engine.tex
\section{Fundus-Engine}

\begin{figure*}[ht]
\centering
\includegraphics[width=17.3cm]{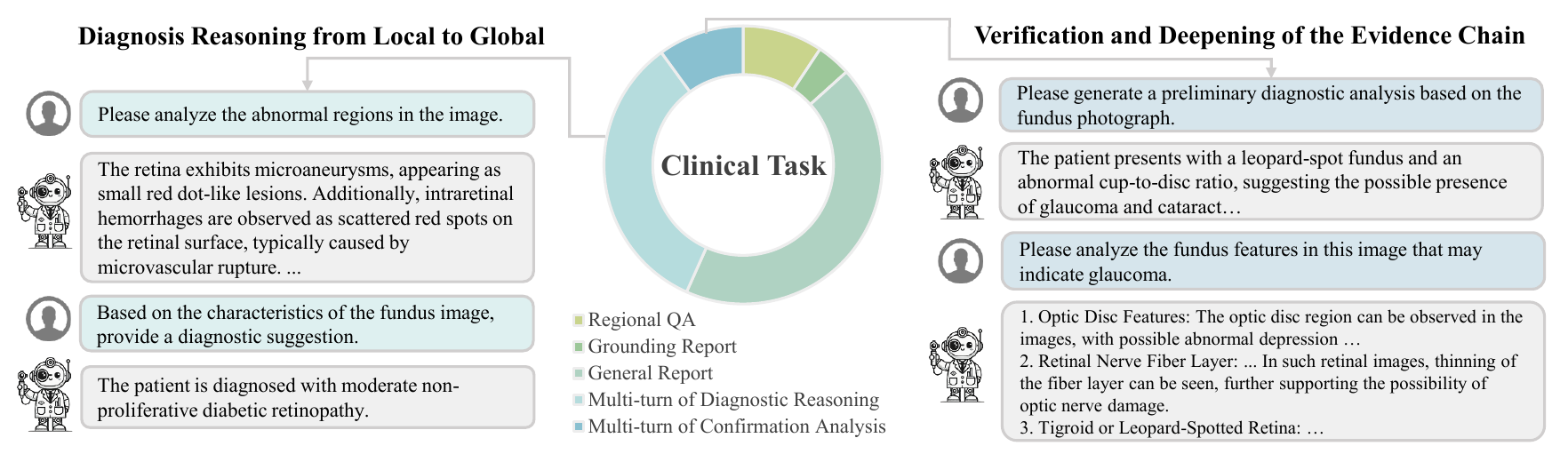}
\caption{Two Typical Interaction Patterns of Clinically Aligned Cognitive Chains and Curated Instruction Fine-tuning Data Scheme.}
\label{fig:task}
\vspace{-2mm}
\end{figure*}

We construct the Fundus-Engine achieving the collaborative annotation of disease classification, regional localization, and lesion characteristics within a single fundus image. 
By simulating the progressive cognitive process of human clinicians during diagnosis, we establish an explicit cognitive chain, resulting in FundusGen.  

The Fundus-Engine system employs a dual-path automated annotation framework to enhance the efficiency:  
(1) The automatic bounding box annotation system utilizes a semi-supervised self-training nnU-Net to perform regional segmentation, followed by a spatial clustering algorithm to aggregate pixel-level predictions into bounding boxes.  
(2) An MLLM-driven semantic expansion algorithm transforms discrete medical annotations into clinically standardized descriptive texts.  

\subsection{Three-Stage Collaborative Annotation}  

\noindent\textbf{Stage 1: Fine-Grained Label Collection.}  
We collect 20w fundus images with fine-grained annotations from open-source datasets (e.g.\cite{wu2024mm,nakayama2024brset,porwal2018indian,aptos2019blindnessdetection, Kovalyk2022, decenciere2014feedback}) and in-house datasets. These images are expert-annotated by ophthalmologists, providing granular labels for both global disease classification and detailed lesion annotations.

\noindent\textbf{Stage 2: Automated Bounding Box Annotation System.}  
To address the lack of bounding box annotations in existing datasets, we propose an automatic bounding box annotation system. Since open-source datasets generally lack direct bounding box annotations, we generate them by clustering segmentation labels. Based on both clinical guidelines and data availability, we select two representative fundus structures (optic cup and optic disc) and three lesion types (hard exudates, microaneurysms, and cotton-wool spots) for bounding box annotation. The automated bounding box annotation system consists of the following steps:  
\begin{itemize}
    \item \textbf{Data Preprocessing:} We integrate segmentation annotations from publicly available datasets and apply the filtering method from \cite{zhou2022automorph} to remove low-quality images, constructing a foundational training set ($<$1,000 samples per category).  
    \item \textbf{Model Training:} For different categories, we build separate nnU-Net \cite{isensee2021nnu} segmentation networks, leveraging a feature pyramid structure to extract multi-scale features.  
    \item \textbf{Semi-Supervised Expansion:} We design an iterative self-training process that generates pseudo-labels for unannotated lesion regions in images labeled during Stage One. These pseudo-labels are incorporated into the training set for iterative optimization. We evaluate the performance of the automated segmentation annotation on cross-domain datasets and demonstrate the feasibility of pseudo-labels through experimental results presented in the appendix.  
    \item \textbf{Bounding Box Generation:} The DBSCAN clustering algorithm \cite{ester1996density} is applied to convert pixel-level segmentation results into standardized bounding box annotations.  
\end{itemize}

\noindent\textbf{Stage 3: MLLM-Driven Semantic Expansion.}  
To bridge the semantic gap between discrete annotations and the training requirements of multimodal models, we employ an MLLM-based cross-modal alignment framework.  
\begin{itemize}
    \item \textbf{Structured Parsing:} The annotation results from Stages One and Two are transformed into structured labels.  
    \item \textbf{Prompt Engineering:} A constraint-driven prompt framework is designed to ensure that the generated text meets the following criteria:  
    (1) \textit{Observational Objectivity}: All descriptions are traceable to image pixel features.  
    (2) \textit{Clinical Relevance}: Implicit diagnostic clues are incorporated while avoiding conclusive statements.  
    \item \textbf{Text Generation:} The structured annotations are mapped to natural language descriptions via the MLLM(GPT-4o), producing compound texts that integrate localization information (e.g., "optic disc located at [box]") and diagnostic reasoning (e.g., "fundus disease grading inferred based on lesion distribution").  
    \item \textbf{Quality Control:} Clinical experts review the generated texts in a double-blind manner to ensure compliance with clinical standards. Texts that do not meet the required quality are discarded or regenerated.
\end{itemize}

\subsection{Construction of Clinically Aligned Cognitive Chains}  

To simulate the progressive cognitive process of human clinicians from lesion observation to comprehensive diagnosis, in the training, we construct cognitive chains within MLLMs using multi-turn dialogues. This approach guides the model in generating interpretable reasoning paths, as shown in Figure \ref{fig:task}. The core logic follows a "\textit{region localization} $\rightarrow$ \textit{feature analysis} $\rightarrow$ \textit{diagnostic reasoning}" cognitive pathway, with typical interaction patterns as follows:  

\subsubsection{Diagnosis Reasoning from Local to Global}  

The first dialogue round focuses on abnormal regions, with a prompt such as "Please analyze the abnormal regions in the image." The model's response includes an analysis of the abnormal areas, incorporating positional information if available. In the second round, the model is prompted to integrate clinical knowledge for analysis, such as "Based on the characteristics of the fundus image, provide a diagnostic suggestion." At this stage, the model may integrate previously mentioned abnormal region characteristics and spatial distributions, obtaining the diagnostic conclusion that "The patient is diagnosed with ..."

\subsubsection{Verification and Deepening of the Evidence Chain}  

The initial instruction prompts the model to generate a coarse-grained diagnostic overview based on image features: "Please generate a preliminary diagnostic analysis based on the fundus image." The model synthesizes the image features to provide an initial assessment:  "...suggesting a potential presence of glaucoma and cataracts...". Subsequent dialogues refine the evidence chain by prompting verification of fine-grained features for specific diseases. For example, a follow-up query may ask: "Please analyze the fundus features in this image that may indicate glaucoma." The model then expands on its reasoning based on image features:  
"1. Optic Disc Features: ..."

%% file: sec/4_fundusexpert.tex
\section{FundusExpert}

FundusExpert is a positioning-diagnosis collaborative multimodal model designed to address visual-language tasks related to ophthalmology. In this section, we first introduce the curated scheme for instruction fine-tuning data. Then, we present the training process of the model. Finally, we introduce Fundus-MMBench, a standardized multimodal evaluation framework focused on fundus images.

\subsection{Curated Instruction Fine-tuning Data Scheme}\label{sec:data_select}

To meet the core needs of clinical ophthalmology, we design instructions for various tasks to enhance the model's profound understanding of fundus images, including:  
(1) \textbf{General Report:} Generate standardized diagnostic reports (e.g., "Generate a diagnostic report based on the fundus image"). This part of the data serves as the initialization data in fine-tuning experiments, and the ablation results are presented in the Experiment section;  
(2) \textbf{Regional QA:} These instruction data can be generated through rules, focusing on localization and identification (e.g., "Label the location of hard exudates");  
(3) \textbf{Grounding Report:} The report content must be directly associated with image regions (e.g., "Describe the fundus image with positional information");  
(4) \textbf{Multi-turn Diagnostic Reasoning:} Simulate the doctor's questioning process, diagnose diseases based on characteristics of abnormal regions and their location distribution (e.g., "Provide a diagnostic suggestion based on the characteristics of the fundus image");  
(5) \textbf{Multi-turn Confirmation Analysis:} Verify and deepen the evidence chain through multi-turn dialogues (e.g., "Describe and analyze the fundus features related to glaucoma in this image"). Instructions (4) and (5) explicitly demonstrate the construction of cognitive chains.

\subsection{FundusExpert Training} \label{sec:fundusexpert_training}

Given the powerful multimodal understanding capabilities of InternVL2.5 \cite{chen2024expanding} and the extensive medical knowledge pre-training, we use it as the base model for instruction fine-tuning. The performance of InternVL2.5, as shown in Table \ref{tb:model_comparison}, surpasses even larger medical domain-specific models like MedDr in terms of its performance on the ophthalmic fundus image modality. We fully fine-tune the entire network on the 8B version to obtain FundusExpert, which has a visual encoder of 300M parameters (InternViT) and a language encoder of 7B parameters (InternLM). Since our training data consists of curated instruction fine-tuning data, the entire training process involves only the instruction fine-tuning. 

\subsection{Fundus-MMBench}
In the objective evaluation of our experiments, we analyze model performance using two standardized evaluation frameworks: GMAI-MMBench \cite{chen2024gmai} and Fundus-MMBench. These two frameworks are forming a complementary evaluation system.

GMAI-MMBench \cite{chen2024gmai}, as a general medical multimodal benchmark, covers various medical imaging modalities, including X-ray, CT, and fundus image. This study focuses on its fundus image subset, which comprehensively includes 60 task categories, primarily involving ophthalmic disease diagnosis and grading, including over ten rare ophthalmic diseases. Notably, approximately 45\% of the disease categories are not explicitly annotated in our training data, making this benchmark effective in assessing FundusExpert’s generalization ability in out-of-distribution scenarios. However, as a general medical evaluation system, GMAI-MMBench exhibits a sample distribution imbalance in the fundus photography modality, where over 90\% of the categories contain only around 5 test samples, potentially leading to significant variability in evaluation results due to small sample effects.

To address the clinical needs of fundus image, we construct Fundus-MMBench, a multimodal evaluation framework dedicated to fundus imaging. Fundus-MMBench increases the number of test samples per task category to 20. It consists of 31 fine-grained tasks covering three core clinical domains: region-based object recognition (e.g., optic disc identification), disease classification (e.g., glaucoma vs. non-glaucoma diagnosis), and severity grading (e.g., diabetic retinopathy severity assessment). Our training dataset FundusGen is strictly isolated from Fundus-MMBench. All evaluation categories in Fundus-MMBench are represented in the training data, enabling the quantification of FundusExpert’s performance boundaries in in-distribution tasks.

%% file: sec/5_experiment.tex
\section{Experiment}
\input{table/model_compare}

\begin{figure*}[ht]
\centering
\includegraphics[width=\textwidth]{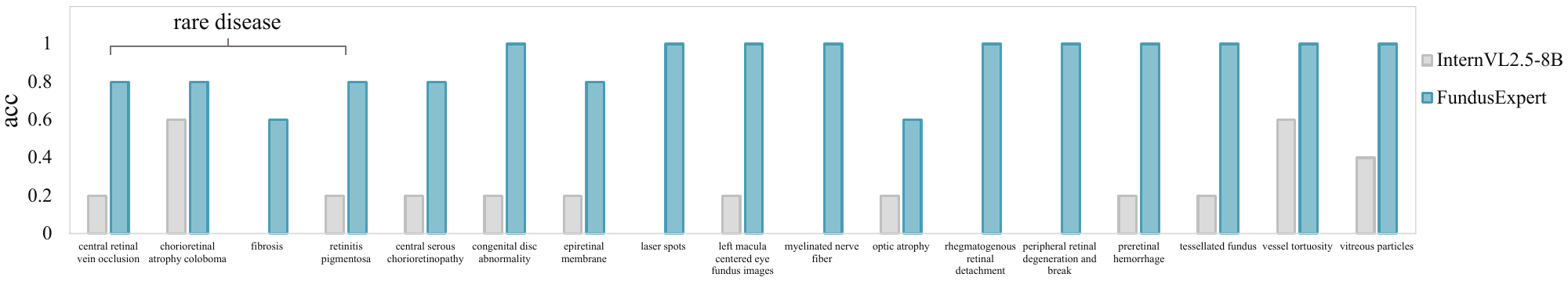}
% \vspace{-0.2cm}
\caption{Zero-shot Performance Improvement for Out-of-domain Testing. (Each category has five samples) }
\label{fig:GMAI_bar_caption}
\vspace{-0.25cm}
\end{figure*}

\subsection{Performance Evaluation}

\subsubsection{Clinical Question Answering Evaluation}
In the clinical QA task, we evaluate the model performance using two objective benchmarks: Fundus-MMBench and GMAI-MMBench \cite{chen2024gmai}(fundus image subset), following the evaluation setup in VLMEvalKit\cite{duan2024vlmevalkit}. We employ a deterministic sampling strategy with a temperature parameter of 0 to ensure response stability. The semantic matching mechanism allows for fault-tolerant answers (e.g., "A. optic cup" being equivalent to option A), making the evaluation more aligned with real-world scenarios. For the evaluation of commercial closed-source models such as GPT-4o, we repeatedly prompt the model until it generates a response, if the model initially refuses to respond.

The evaluation results are shown in Table \ref{tb:model_comparison}. The experimental results indicate that FundusExpert achieves optimal accuracy in both evaluation frameworks, surpassing the ophthalmic vision-language foundation model RetiZero \cite{wang2024common} by an average of 17.6\%. RetiZero \cite{wang2024common} employs a purely contrastive learning framework (CLIP-based), with a large collection of fundus images and pure text data covering over 400 diseases. While it outperforms other models in the GMAI-MMBench evaluation, its pretraining task focuses on zero-shot classification and cross-domain recognition. It can only perform global text matching and cannot directly generate texts.

\noindent\textbf{Validation of the FundusGen Dataset Effectiveness.}  The experiments (Table \ref{tb:model_comparison}) show that when using a unified instruction fine-tuning scheme, multiple Vision-Language models achieve significant performance improvements. After fine-tuning with FundusGen, LLaVA-v1.5 \cite{liu2024improved} shows a 13.1\% average accuracy improvement (26.9\% $\to$ 40.0\%), and Qwen2-VL \cite{wang2024qwen2} shows a 22.2\% average accuracy improvement (34.7\% $\to$ 56.9\%), validating the dataset's value in adapting general multimodal models to the domain.

\noindent\textbf{Extrapolation Ability of FundusExpert.}  
FundusExpert demonstrates significant extrapolation reasoning ability in out-of-domain tasks on GMAI-MMBench. As shown in Table \ref{tb:model_comparison}, it achieves a 66.7\% accuracy rate in zero-shot tasks on GMAI-MMBench, surpassing the base model InternVL2.5 by 30.2\%. This is primarily attributed to FundusGen's explicit modeling of clinical feature inference logic. Case comparisons in Figure \ref{fig:GMAI_bar_caption} further validate this ability.

\subsubsection{Zero-shot Ability in Open-domain Tasks}
\label{sec:Zero-shot Ability}
\noindent\textbf{Localization Ability Evaluation.} 
For the localization boxes output by the model, we perform a quantitative analysis using Intersection over Union (IoU). The spatial alignment at the pixel level is used to assess the localization accuracy. The zero-shot results are shown in Table~\ref{tb:iou}.

\input{table/iou}

\noindent\textbf{Clinical Consistency Evaluation in Medical Report Generation.} 
Existing likelihood-based benchmarks for medical text generation, such as BLEU and ROUGE, inadequately assess semantic plausibility. To overcome this, we introduce a multi-granularity semantic matching framework that evaluates the accuracy of generated medical reports. This framework leverages a VLM(GPT-4o), to perform a structured evaluation of clinical logical consistency.

Let the set of ground-truth labels be $\mathcal{L} = \{l_1, l_2, ..., l_N\}$, which includes both positive and negative findings. Let the set of semantic features extracted from the generated report be $\mathcal{S} = \{s_1, s_2, ..., s_M\}$. The clinical consistency score is defined as:
\vspace{-1mm}
\[
\text{Clinical Consistency} = \frac{\sum_{i=1}^N \mathbb{I}(\text{match}(l_i, \mathcal{S}))}{|\mathcal{L} \cup \mathcal{S}|}
\]
where, the function $\text{match}(l_i, \mathcal{S})$ checks for a bidirectional semantic correspondence between a label $l_i$ and the set of generated features $\mathcal{S}$, as determined by the VLM. $\mathbb{I}(\cdot)$ is the indicator function, which is 1 if the condition is true and 0 otherwise. The denominator $|\mathcal{L} \cup \mathcal{S}|$ is the size of the union of the ground-truth labels and the generated features(determined by the VLM), which normalizes the score.

In the evaluation of 200 reports generation, FundusExpert achieves 77.0\% in clinical consistency, significantly outperforming GPT-4o, which scores 47.6\% (+29.4\%).

\subsection{Verification of Dataset Scaling Law}

\begin{figure}[htbp]
    \centering
    \begin{minipage}{0.48\textwidth}
        \centering
        \includegraphics[width=0.49\textwidth]{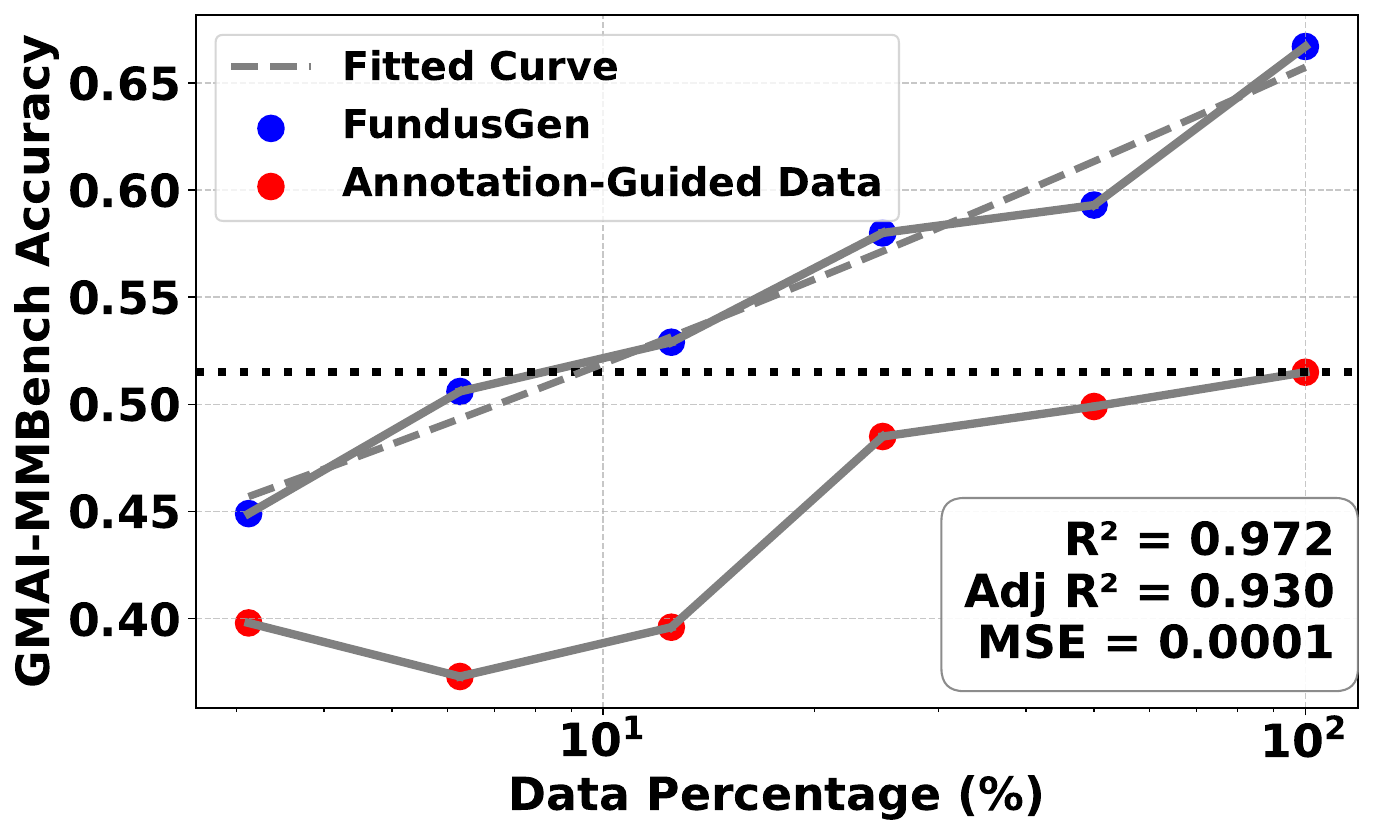}
        \includegraphics[width=0.49\textwidth]{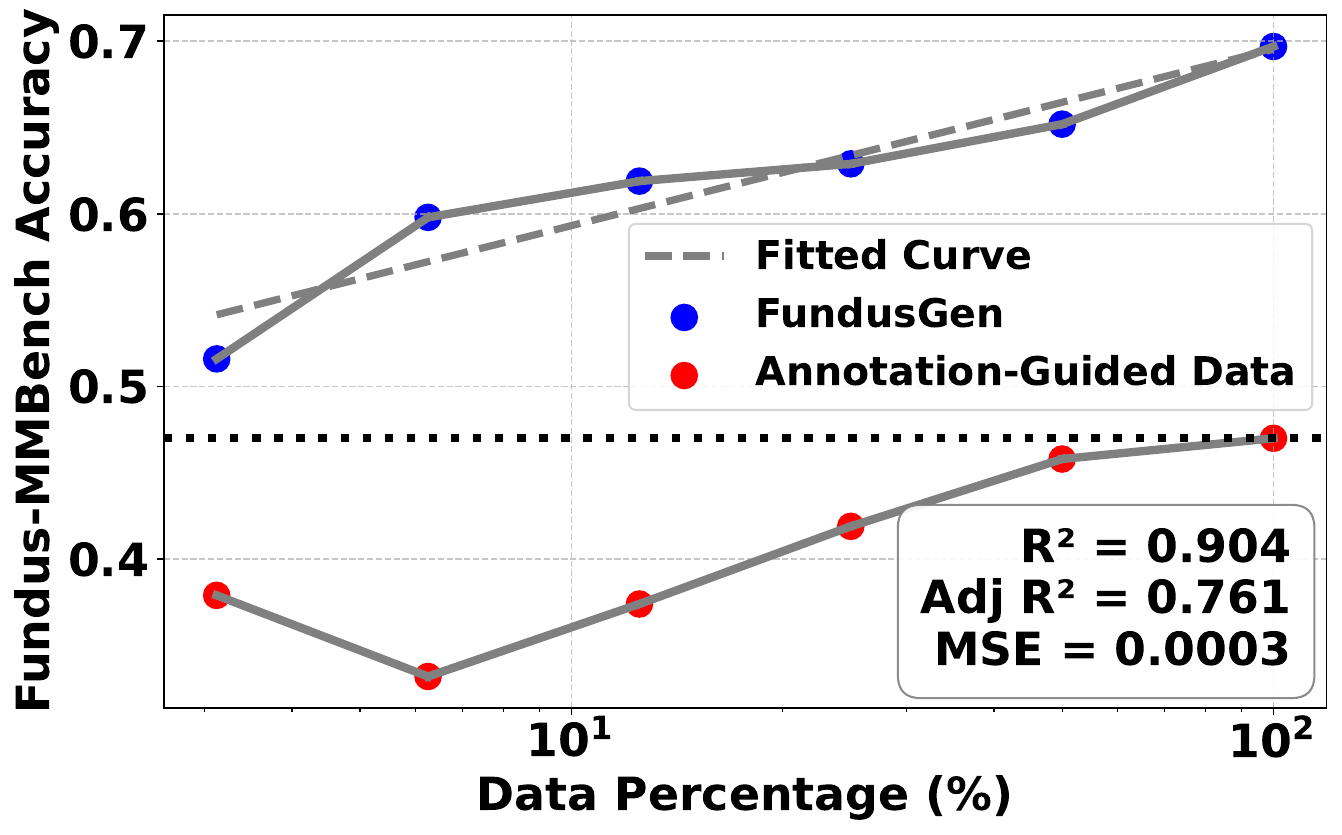}
        \caption{The change of model accuracy on GMAI-MMBench(left) and Fundus-MMBench(right) under different data percentage. The blue dots represent the model trained with FundusGen, and the red dots represent the model trained with Classification Annotation-Guided Data. The gray dotted line is the fitting curve, showing scaling law.}
        \label{fig:scaling_law_plot}
    \end{minipage}
    
\end{figure}

In the subset sampling experiment of the FundusGen dataset (fine-tuning based on InternVL2.5-8B), the model trained with FundusGen demonstrates a scaling law with respect to the percentage of data, as shown in Figure \ref{fig:scaling_law_plot}. This indicates that the FundusGen dataset, built on clinical cognitive chains, possesses high information density, strong scalability, and low semantic noise outperforming comparison datasets such as Classification Annotation-Guided Data. The latter is generated by directly combining fine-grained annotations(the same disease labels and feature labels as FundusGen) with images and using GPT-4o to produce comparative data. However, Classification Annotation-Guided Data does not have localization information and lacks the explicit instructions necessary to guide the model in generating interpretable reasoning paths.

For performance in GMAI-MMBench, The fitted curve shows that models trained with FundusGen (blue points) exhibit a significant power-law scaling behavior, which can be expressed as \( L \propto N^{\alpha}, \alpha = 0.068 \), with a correlation coefficient of \( R^2 = 0.972 \), adjusted \( R^2 = 0.930 \), and MSE = 0.0001. As the data volume increases, the model trained with FundusGen diagnostic accuracy shows a stable and predictable upward trend. In contrast, models trained with Classification Annotation-Guided Data (red points) show no significant performance improvement as the data size increases, and in some data intervals, performance even declines. For performance in GMAI-MMBench, a comparative experiment reveals that fine-tuning on a 10\% subset of FundusGen produces results comparable to those obtained using 100\% of Classification Annotation-Guided Data.

\subsection{Ablation on Dataset}
The ablation experiments (Table \ref{tb:ablation}) aim to validate the effectiveness of the data selection scheme in Section \ref{sec:data_select}, with the experimental setup being the same as in Section \ref{sec:fundusexpert_training}.

\input{table/ablation}

\noindent\textbf{Cognitive Chain Construction Data Ablation.}
To validate the effectiveness of the explicit reasoning mechanism in multi-turn dialogues, we compare: (1) Full FundusGen group: directly using FundusGen; (2) Cognitive Chain Degradation group: splitting multi-turn diagnostic instructions into independent single-turn tasks, which disrupt the continuity of the cognitive chain, with random sampling of instructions of the same scale as (1). The results of the clinical QA task evaluation are shown in Table \ref{tb:ablation}. The average diagnostic accuracy for diseases in the GMAI-MMBench decreases by 3.5\% for the (2) Cognitive Chain Degradation group compared to (1), indicating that reasoning by constructing a progressive chain, enhances the model's logical deduction ability for complex pathologies.

\input{table/messidor_compare}

\noindent\textbf{Region-Aware Data Ablation.}
To verify the effectiveness of the positioning-diagnosis coordination mechanism, we compare: (3) FundusGen Control group 1: randomly sampling from FundusGen with the same scale of instructions as (4); (4) Region Data Removal group: removing all instructions in FundusGen containing region annotations (e.g., bounding box annotations, region-text alignment tasks) while retaining other types of tasks. The experimental results show that the (4) Region Data Removal group experiences an overall decrease of 3.6\% in Fundus-MMBench. Additionally, group (4) experiences an overall decline of 5.4\% on the out-of-distribution GMAI-MMBench, indicating that region-level annotations contribute to the model's spatial semantic understanding and enhance its generalization ability.

\noindent\textbf{Startup Data Ablation.}
To quantify the value of using General Report(enhances the model's basic understanding of different diseases) as startup data, we compare: (5) FundusGen Control group 2: randomly sampling from FundusGen with the same scale of instructions as (6); (6) No Startup Data group: removing all standardized diagnostic report data. In addition to the performance degradation in Table \ref{tb:ablation} (training for 1 epoch), further experiments show that (6) requires 0.5 additional epochs (training for 1.5 epochs) to achieve the same accuracy as (5) on Fundus-MMBench, indicating that there is a delay in convergence without startup data.

% At the same time, its performance on out-of-distribution GMAI-MMBench worsens.

\subsection{Data Generator}
\input{table/data_generation}

Experiment results(\ref{sec:Zero-shot Ability}) from clinical consistency evaluation show that FundusExpert outperforms GPT-4o in zero-shot ophthalmic medical report generation and domain knowledge comprehension. This finding suggests a paradigm shift in data generation from commercial closed-source models (e.g., GPT-4o) to lightweight domain-specific models, enabling more efficient iterative data acquisition. 
% Synthetic data mitigates privacy concerns associated with real patient records while also reducing annotation costs.

We replace GPT-4o with FundusExpert as the MLLM for semantic expansion in Fundus-Engine. By leveraging the multi-level annotation labels in the dataset, we generate medical texts for 100K images(outside the training data) and construct single-turn General Report. Instruction fine-tuning experiments are subsequently conducted on Qwen2-VL\cite{wang2024qwen2} and InternVL2.5\cite{chen2024expanding}. As shown in Table~\ref{tb:data_gen}, the fine-tuning results using synthetic data from the lightweight domain-specific model surpass those obtained from GPT-4o-generated data.

\subsection{Pseudo-Label Accuracy Evaluation}  

This experiment aims to assess the quality and usability of the bounding box pseudo-labels(Table \ref{tab:messidor_compare}). 
% When training the nnU-Net-based segmentation model using open-source data, we do not include the Messidor fundus image dataset\cite{decenciere2014feedback}. 
In this experiment, Messidor\cite{decenciere2014feedback} serves as an out-of-domain(OOD) test set to evaluate the cross-domain predictive performance of models trained with real labels versus pseudo-labels. Our primary objective is to demonstrate the segmentation model's performance in OOD tasks, as the process of annotating in-house data with bounding box pseudo-labels using a model trained on open-source data inherently represents a OOD task.

%% file: table/model_compare.tex
\begin{table*}[ht]
    \centering
    \small
    \setlength{\tabcolsep}{14pt}
    \renewcommand{\arraystretch}{0.99} 
    \begin{threeparttable}
    \begin{tabular}{cccccc}
        \toprule[0.5mm]
        \makecell[c]{\textbf{Type}} & \makecell[c]{\textbf{Model}} & \makecell[c]{\textbf{Params}} & \makecell[c]{Fundus-MMBench} & \makecell[c]{GMAI-MMBench\cite{chen2024gmai}$^\dag$} \\
        \midrule[0.3mm]
        \multirow{4}{*}{Specialist} 
            & MedDr\cite{he2024meddr} & 40B & 34.8\% & 33.7\% \\
            & MedRegA\cite{wang2024interpretable} & 40B & 40.3\% & 40.1\% \\
            & GMAI-VL\cite{li2024gmai} & 8B & 44.5\% & 56.1\% \\
             & RetiZero\cite{wang2024common}$^\ddag$ & - & 42.0\% & 59.2\% \\
        \midrule[0.3mm]
        \multirow{5}{*}{Generalist} 
            & LLaVA-v1.5\cite{liu2024improved} & 7B & 21.1\% & 32.7\% \\
            & Qwen2-VL\cite{wang2024qwen2} & 7B & 33.5\% & 35.9\% \\  
            & Qwen2.5-VL\cite{bai2025qwen2} & 7B & 30.6\% & 37.8\% \\
            & InternVL2.5\cite{chen2024expanding} & 8B & 40.6\% & 36.5\% \\
            & GPT-4o\cite{achiam2023gpt} & - & 41.6\% & 57.4\% \\
            & Gemini-2.0-pro\cite{team2023gemini} & - & 46.1\% & 59.0\% \\
        \midrule[0.3mm]
        \multirow{4}{*}{Fine-tuned by FundusGen} 
            & LLaVA-v1.5* & 7B & 38.2\%(+17.1\%) & 41.7\%(+9.0\%) \\
            & Qwen2-VL* & 7B & 57.4\%(+23.9\%) & 56.4\%(+20.5\%) \\
            & FundusExpert-mini & 1B & 63.5\%(+30.0\%) & 58.3\%(+28.5\%) \\
            & \textbf{FundusExpert} & 8B & \textbf{69.7\%}(+29.1\%) & \textbf{66.7\%}(+30.2\%) \\
        \bottomrule[0.5mm]
    \end{tabular}
    \begin{tablenotes}
    \item[$^\dag$] Only fundus images are selected for evaluation. GMAI-MMBench mentioned in the following text refers to the same subset.
    \item[$^\ddag$] CLIP-base vision-language foundation model.
    \item[*] Fine-tuned by FundusGen.
    \end{tablenotes}
    \end{threeparttable}
    \vspace{-1mm}
    \caption{Performance comparison on clinical QA tasks.}
    \label{tb:model_comparison}
\end{table*}

%% file: table/iou.tex
\begin{table}[h]
    \centering
    \footnotesize
    \renewcommand{\arraystretch}{0.99} 
    \setlength{\tabcolsep}{2.8pt}  % column space
    % \rowcolors{2}{gray!15}{white} 
    \begin{threeparttable}
    \begin{tabular}{cccccc}
        \toprule[0.35mm]
        \makecell[c]{\textbf{Model}} & \makecell[c]{$IoU_{OC}$} & \makecell[c]{$IoU_{OD}$} & \makecell[c]{$IoU_{EX}$} & \makecell[c]{$IoU_{CWS}$} & \makecell[c]{$IoU_{MA}$}   \\
        \midrule[0.2mm]
        LLaVA-1.5-7B\cite{liu2024improved}&  0.021 & 0.064 & 0.005 & 0.002 & 0.002  \\
        InternVL2.5-8B\cite{chen2024expanding}& 0.036  &  0.077 &  0.035 & 0.004 & 0.007  \\
        MedRegA\cite{wang2024interpretable}  & 0.302 & 0.543 & 0.038 & 0.006 & 0.011 \\
        \textbf{FundusExpert}  & \textbf{0.632} & \textbf{0.738} & \textbf{0.194} & \textbf{0.141} & \textbf{0.116} \\

        \bottomrule[0.35mm]
    \end{tabular}

    \begin{tablenotes}[flushleft]
    \footnotesize
    \item \textit{Note:} OC: Optic Cup, OD: Optic Disc, EX: Hard Exudates, CWS: Cotton Wool Spots, and MA: Microaneurysms.

    \end{tablenotes}
    \end{threeparttable}
    
    \caption{Performance on the object detection task.}
    \label{tb:iou}
\vspace{-3mm}
\end{table}

% Optic Cup   Optic Disc  Hard Exudates  Cotton Wool Spots   Microaneurysms

%% file: table/ablation.tex
\begin{table}[ht]
    \centering
    \scriptsize
    \setlength{\tabcolsep}{1.8pt}
    \begin{tabular}{cccc}
        \toprule[0.35mm]
        \textbf{Data} & \textbf{Modification} & \textbf{Fundus-MMBench} & \textbf{GMAI-MMBench} \\
        \midrule[0.2mm]
        % First group of experiments (Cognitive chain construction)
        (1) Complete FundusGen & \multirow{2}{*}{\shortstack{Cognitive \\ Chain}} & 69.7\% & 66.7\% \\
        (2) Cognitive Degradation & & 67.1\% ($\downarrow$2.6\%) & 63.2\% ($\downarrow$3.5\%) \\
        \midrule[0.2mm]
        % Second group of experiments (Region perception data)
        (3) Sampled FundusGen 1 & \multirow{2}{*}{\shortstack{Region \\ Perception}} & 68.9\% & 64.7\% \\
        (4) Region Data Removal & & 65.3\% ($\downarrow$3.6\%) & 59.3\% ($\downarrow$5.4\%) \\
        \midrule[0.2mm]
        % Third group of experiments (Initialization data)
        (5) Sampled FundusGen 2 & \multirow{2}{*}{\shortstack{Startup \\ Data}} & 66.2\% & 60.9\% \\
        (6) Startup Data Removal & & 62.9\% ($\downarrow$3.3\%) & 56.4\% ($\downarrow$ 4.5\%) \\
        \bottomrule[0.35mm]
    \end{tabular}
    \caption{Ablation Experiment. Comparison of Clinical Question Answering Results under Different Data Conditions and the Same Training Setup.}
    \label{tb:ablation}
\end{table}

%% file: table/messidor_compare.tex
\begin{table*}[ht]
    \renewcommand{\arraystretch}{0.97} 
    \setlength{\tabcolsep}{2pt} 
    \caption{Out-of-domain data test results. The same number of image-true labels and image-pseudo labels of different categories are used as training data to train different segmentation models and test them on Messidor\cite{decenciere2014feedback}.}
    \label{tab:messidor_compare}
    \centering
    \small
    \begin{threeparttable}
    \begin{tabular*}{\textwidth}{@{\extracolsep{\fill}} ccccccc}
        \toprule[0.4mm]
        \multirow{2}{*}{\textbf{Category}} & \multicolumn{2}{c}{\textbf{True Labels}} & \multicolumn{2}{c}{\textbf{Pseudo Labels\textsubscript{1st\_round\_prediction}}} & \multicolumn{2}{c}{\textbf{Iterative Pseudo Labels\textsubscript{2nd\_round\_prediction}}} \\
        \cmidrule(lr){2-3} \cmidrule(lr){4-5} \cmidrule(lr){6-7}
        & \makecell[c]{$Dice$} & \makecell[c]{$IoU_{pixel\_level}$} & \makecell[c]{$Dice$} & \makecell[c]{$IoU_{pixel\_level}$} & \makecell[c]{$Dice$} & \makecell[c]{$IoU_{pixel\_level}$} \\
        \midrule[0.25mm]
        Hard Exudates  & 29.1\% & 19.6\% & 29.9\% & 20.3\% & 29.6\% & 20.2\% \\
        Microaneurysms  & 21.1\% & 12.4\% & 21.3\% & 12.7\% & 25.0\% & 15.2\% \\
        Cotton-wool Spots  & 28.3\% & 20.3\% & 25.9\% & 18.3\% & 28.6\% & 20.2\% \\
        Optic Cup  & 57.3\% & 45.9\% & 61.4\% & 50.3\% & 60.1\% & 49.4\% \\
        Optic Disc  & 75.3\% & 66.5\% & 82.0\% & 73.4\% & 80.0\% & 71.2\% \\
        \bottomrule[0.4mm]
    \end{tabular*}
    
    % \begin{tablenotes}
    % \item[1] IoU is calculated based on pixel overlap between predictions and true segmentations.
    % \item[*] Results using true labels for training.
    
    % \item[$^\dag$] Results using pseudo labels from the first round of direct segmentation predictions.
    % \item[$^\ddag$] Results using pseudo labels after an additional iterative self-training round.
    % \end{tablenotes}
    
    \end{threeparttable}
    \vspace{-2mm}
\end{table*}

%% file: table/data_generation.tex
\begin{table}[ht]
    \centering
    \small
    \renewcommand{\arraystretch}{0.74} 
    % \setlength{\tabcolsep}{2.8pt}
    % \rowcolors{2}{gray!15}{white}
    \begin{threeparttable}
    \begin{tabular}{ccc}
        \toprule[0.35mm]
        \makecell[c]{\textbf{Model}} & \makecell[c]{Fundus-MMBench} & \makecell[c]{GMAI-MMBench} \\
        \midrule[0.2mm]
        Qwen2-VL-7B$^\dag$  & 35.6\% & 39.4\% \\
        Qwen2-VL-7B* & 40.1\% & 41.7\% \\
        InternVL2.5-8B$^\dag$  & 48.2\% & 46.2\% \\
        InternVL2.5-8B* & 50.8\% & 48.4\% \\
        \bottomrule[0.35mm]
    \end{tabular}
        \begin{tablenotes}
    \item[$^\dag$] Fine-tuned by GPT-4o-generated data.
    \item[*] Fine-tuned by FundusExpert-generated data.
    \end{tablenotes}
    \end{threeparttable}
    
    \caption{Comparison of synthetic data fine-tuning performance.}
    \label{tb:data_gen}
    \vspace{-2mm}
\end{table}

%% file: sec/6_conlcusion.tex
\section{Conclusion and Discussion}
We present FundusExpert, an ophthalmic MLLM with integrated localization-diagnosis reasoning, along with the FundusGen dataset featuring hierarchical semantic fusion. Our experiments show that FundusExpert achieves 69.7\% diagnostic accuracy on the Fundus-MMBench exceeding GPT-4o by 28.1\% and attains 98\% accuracy in regional identification tasks. We also reveal scaling law in medical multimodal data ($L \propto N^{0.068}$) and investigate that cognitive-aligned annotations enhance data utilization efficiency.

The performance of FundusExpert provides a foundation for the next stage of development, which involves broadening its adaptability for dynamic reasoning and diverse scenarios. Promising methodologies, including test-time scaling and reinforcement learning-based post-training like Deepseek-r1\cite{guo2025deepseek}, can guide this evolution. Future research will integrate reinforcement learning with FundusGen's semantic hierarchy to enhance model adaptability in low-annotation settings, and expand its performance envelope in specialized medical applications.

%% file: sec/7_acknowledgment.tex
\section{Acknowledgment}

This work was supported by Shanghai Artificial Intelligence Laboratory. We would like to thank Professor Yu Qiao for his support for this project. We also thank Professor Xiaodong Sun, Professor Huixun Jia, Yanlin Qu, and Guanran Zhang from Shanghai General Hospital, as well as Zefeng Yang from the Zhongshan Ophthalmic Center of Sun Yat-sen University, for their contributions in labeling and reviewing the experimental data. We are also grateful to Guang Liang for completing parts of the paper writing.

%% file: sec/supplementary.tex
% \clearpage
\setcounter{page}{1}
% \maketitlesupplementary
\setcounter{section}{0}
\setcounter{table}{0}
\setcounter{figure}{0}
\renewcommand{\thesection}{\Roman{section}}
\renewcommand{\thetable}{S\arabic{table}}
\renewcommand{\thefigure}{S\arabic{figure}}

% \vspace{3mm}

\begin{flushleft}
    \LARGE
    \textbf{Appendix}
\end{flushleft}

\section{FundusGen Details}

\subsection{Data Sources and Annotation}

We collect approximately 200K fundus images and their corresponding annotations from both open-source datasets and in-house data.

\noindent\textbf{MM-Retinal\cite{wu2024mm}.} MM-Retinal is a multimodal dataset comprising high-quality image-text pairs collected from professional ophthalmology textbooks.

\noindent\textbf{BRSET\cite{nakayama2024brset}.} BRSET is the first Brazilian multi-label ophthalmic dataset. It consists of retinal fundus photographs centered on the macula, providing extensive global diagnostic disease labels.

\noindent\textbf{IDRiD\cite{porwal2018indian}.} IDRiD is the first dataset representing the Indian population. It includes pixel-level annotations for typical diabetic retinopathy lesions and normal retinal structures. The dataset provides severity grading for diabetic retinopathy and diabetic macular edema for each image.

\noindent\textbf{APTOS2019\cite{aptos2019blindnessdetection}.} This dataset focuses on the severity grading of diabetic retinopathy.

\noindent\textbf{MESSIDOR2\cite{decenciere2014feedback}.} The Messidor-2 dataset is a collection for diabetic retinopathy (DR) screening, where each examination consists of two macula-centered fundus images, one for each eye.

\noindent\textbf{PAPILA\cite{Kovalyk2022}.} This dataset contains medical records and binocular fundus images from the same patient. It also provides segmentation annotations for the optic cup and optic disc, along with patient-level labels based on clinical assessment.

\noindent\textbf{Retina\cite{jr2ngb2025cataractdataset}.} This dataset consists of normal and cataract fundus images for cataract detection.

\noindent\textbf{Glaucoma\_fundus\cite{DVN/1YRRAC_2018}.} This dataset includes glaucoma annotations, providing grading labels for different stages of glaucoma.

\noindent\textbf{In-house Data.} A collection of high-quality color fundus images annotated by professional ophthalmologists, including comprehensive annotations of overall disease diagnoses and characteristic lesions.

\subsection{Curated Instruction Fine-tuning Data Scheme}

This section provides an expanded description of FundusGen. FundusGen is developed to overcome the limitations of conventional ophthalmic datasets and to support the development of domain-specific multimodal large language models (MLLMs) with enhanced clinical reasoning capabilities.

In addition to the annotation process, we curate instruction fine-tuning data tailored to the diverse needs of ophthalmic clinical tasks. We design different types of instructional prompts based on clinical task formats and semantic emphasis:
\begin{enumerate}
    \item \textbf{General Report:} Instructions to generate standardized diagnostic reports (e.g., "Generate a diagnostic report based on the fundus image"). This data serves as startup data during fine-tuning.
    \item \textbf{Regional QA:} Rule-based instructions that focus on localization and identification tasks (e.g., "Label the location of hard exudates").
    \item \textbf{Grounding Report:} Prompts that require the report content to directly correspond to image regions (e.g., "Describe the fundus image with reference to its location information").
    \item \textbf{Multi-turn Diagnostic Reasoning:} Simulated multi-turn dialogues that mimic the clinical inquiry process, where the model integrates information from abnormal regions to generate diagnostic conclusions (e.g., "Based on the characteristics of the fundus image, provide a diagnostic conclusion").
    \item \textbf{Multi-turn Confirmation Analysis:} Multi-turn dialogues that verify and deepen the evidence chain (e.g., "Describe and analyze the fundus features indicative of glaucoma in this image").
\end{enumerate}

Tasks (4) and (5) explicitly construct the cognitive chain, ensuring that the dataset not only covers high-incidence conditions such as diabetic retinopathy and macular edema but also addresses complex diseases like hypertensive retinopathy and age-related macular degeneration.

\begin{figure*}[th]
\centering
\includegraphics[width=0.99\textwidth]{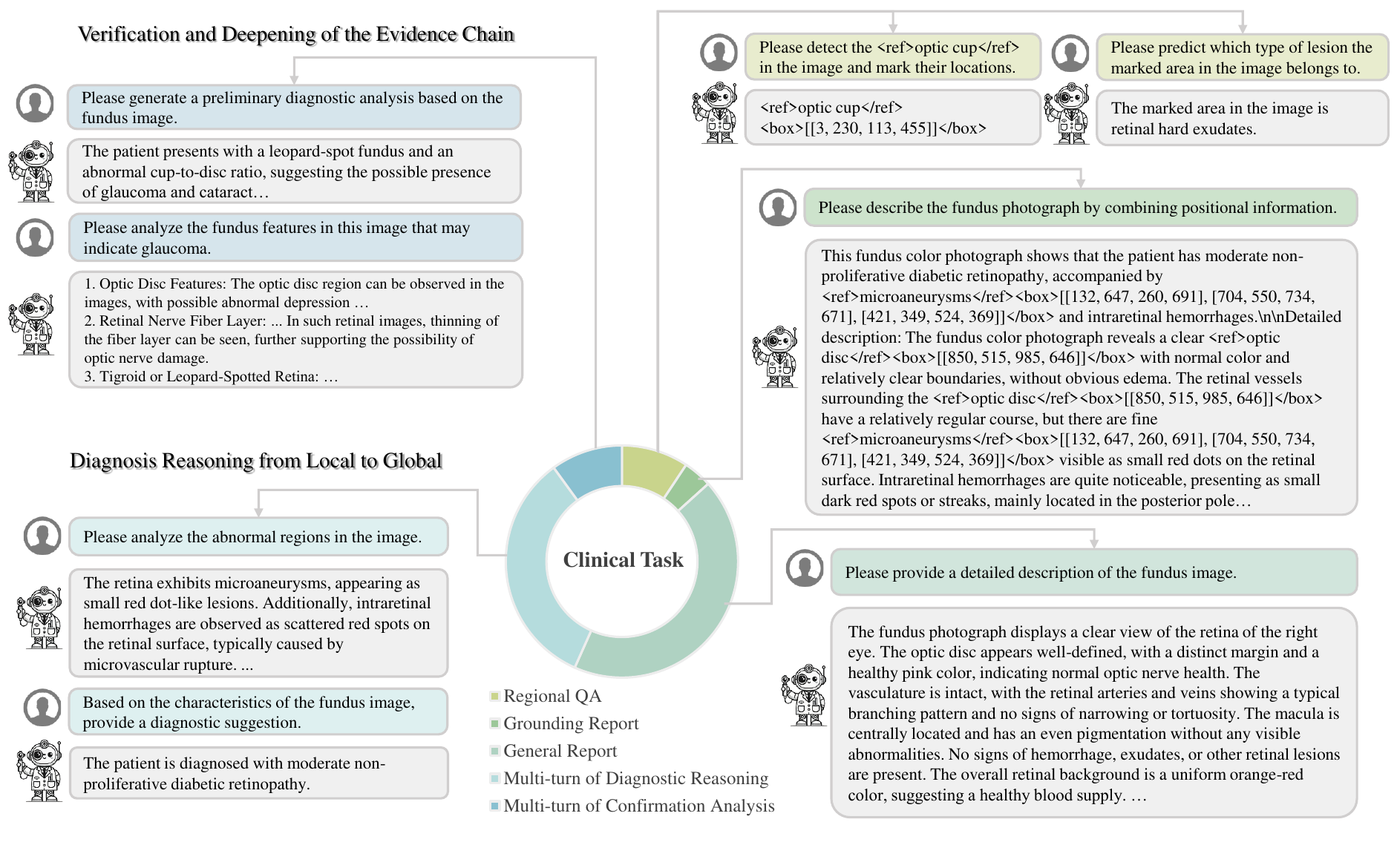}
\vspace{-3mm}
\caption{Data types in FundusGen.}
\label{fig:data_demo}
\end{figure*}

\begin{figure*}[th]
\centering
\includegraphics[width=0.99\textwidth]{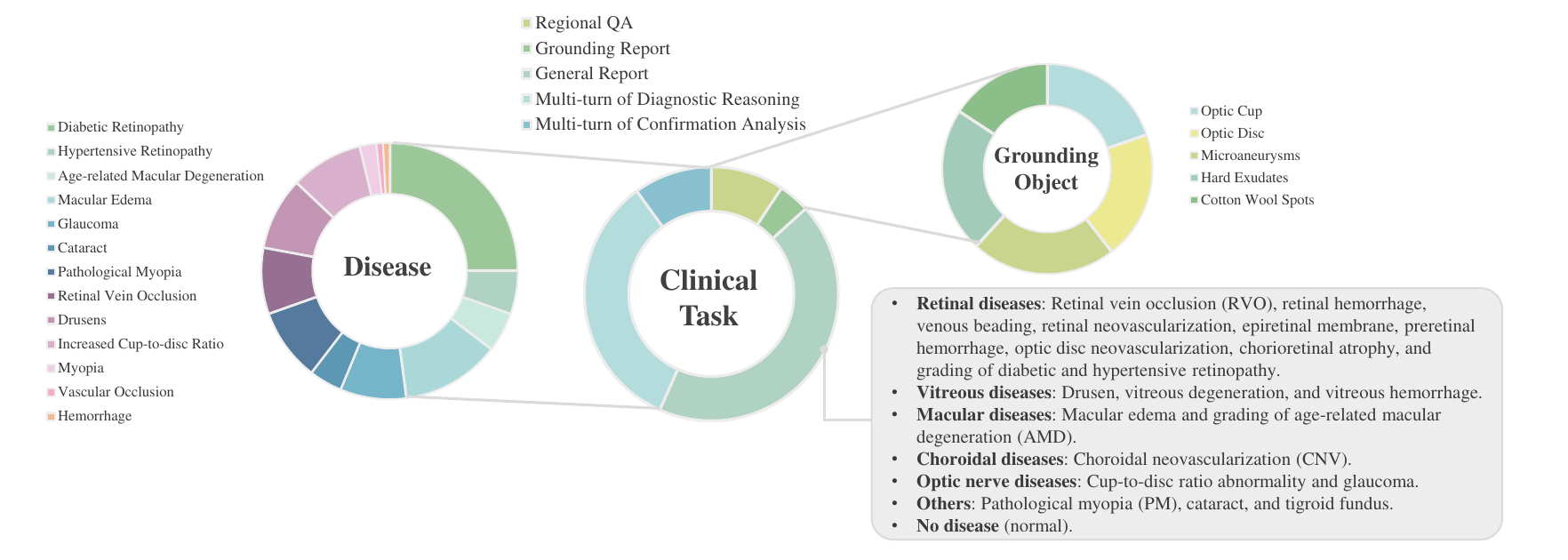}
\vspace{-1mm}
\caption{The Composition of FundusGen.}
\label{fig:task_distribution}
\vspace{-2mm}
\end{figure*}

\section{Fundus-MMBench}
To address the clinical requirements for fundus photography, we construct a multimodal evaluation framework specifically dedicated to fundus images, termed Fundus-MMBench. In Fundus-MMBench, each task category comprises 20 test samples. It consists of 31 fine-grained tasks covering three core clinical domains: region-based object recognition (e.g., optic disc identification), disease classification (e.g., glaucoma versus non-glaucoma diagnosis), and severity grading (e.g., assessment of diabetic retinopathy severity). In the disease classification tasks, we implement a case-control balancing strategy to ensure that the number of positive samples is equal to that of negative samples for each disease, thereby mitigating the impact of data distribution bias on evaluation results. Our training data are strictly isolated from Fundus-MMBench, and all evaluation categories in Fundus-MMBench are represented in the training data, allowing us to quantify the performance boundaries of FundusExpert on in-distribution tasks.

\begin{figure*}[h]
\centering
\includegraphics[width=\textwidth]{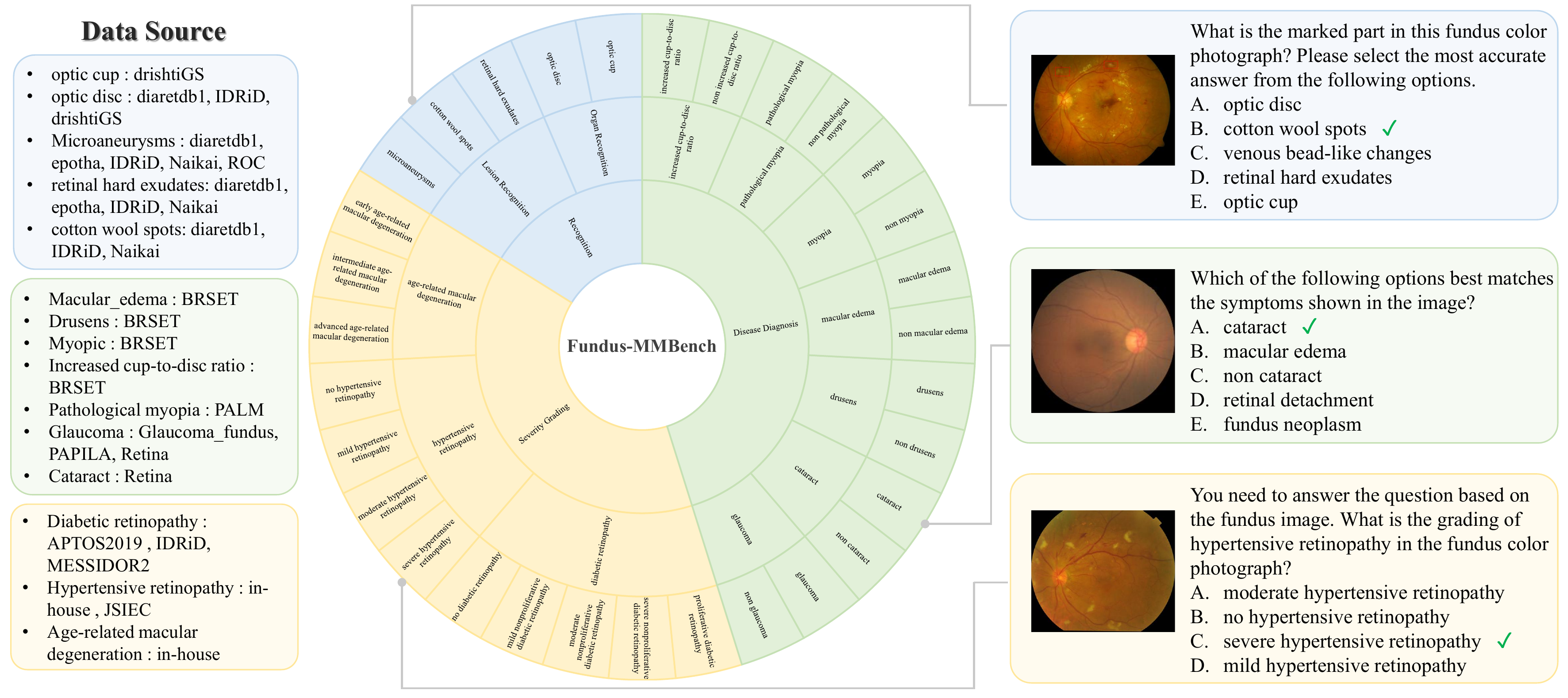}
\caption{The Composition and Presentation of Fundus-MMBench.}
\label{fig:benchmark_detail}
\end{figure*}

\input{table/preudo-label-num}

Given the pervasive issue of class imbalance in medical data—especially where abnormal samples far outnumber normal ones, leading to suboptimal model performance on normal samples and an increased risk of misdiagnosis—we implement a sample balancing strategy in the disease diagnosis evaluation on Fundus-MMBench. For each disease, the number of samples exhibiting the condition is maintained at parity with the number of samples not exhibiting it. By balancing positive and negative samples, we aim to preserve robust disease detection performance while enhancing the model's ability to recognize normal cases, thereby reducing the false positive rate and the risk of misdiagnosis in clinical applications.

\section{Training Details}

\noindent\textbf{Implementation Details For FundusExpert.}  
We employ InternVL2.5\cite{chen2024expanding} as the base model for full-scale instruction tuning. Its vision encoder consists of a 300M InternViT, while its language encoder is a 7B InternLM. During instruction tuning, we unfreeze the vision encoder, MLP, and LLM, optimizing the entire model using 300,000 samples from FundusGen. Training is conducted on four NVIDIA A100 GPUs, with fine-tuning hyperparameters following the official InternVL settings. The per-device batch size is set to 4, with a gradient accumulation step of 8. A cosine learning rate schedule is used, starting at 4e-5, for training over one epoch. We utilize DeepSpeed ZeRO Stage 2 optimization for efficient training.

\noindent\textbf{Implementation Details of Other Architectures.}  
For the fine-tuning of LLaVA-v1.5\cite{liu2024improved} and Qwen2VL\cite{wang2024qwen2}, we adhere to the official InternVL hyperparameter settings, conducting training on four NVIDIA A100 GPUs.

\section{Automated Methods in Fundus-Engine}

\subsection{Bounding Box Generation}  
We apply the DBSCAN clustering algorithm\cite{ester1996density} to convert pixel-level segmentation labels into bounding box annotations (Table \ref{tab:Pseudo_nums}). The epsilon value for DBSCAN clustering is set to 160, and the minimum samples parameter is set to 10. If a bounding box has a pixel area greater than the threshold ($>$100), it is added to the candidate list. The bounding boxes are then sorted by area, and the top three largest bounding boxes are retained.

% \subsection{Pseudo-Label Accuracy Evaluation}  

% This experiment aims to assess the quality and usability of the bounding box pseudo-labels.  

% When training the nnU-Net-based segmentation model using open-source data, we do not include the Messidor fundus image dataset\cite{decenciere2014feedback}. In this experiment, Messidor serves as an out-of-domain test set to evaluate the cross-domain predictive performance of models trained with real labels versus pseudo-labels. Our primary objective is to demonstrate the segmentation model's performance in cross-domain tasks, as the process of annotating in-house data with bounding box pseudo-labels using a model trained on open-source data inherently represents a cross-domain task.

% \input{table/messidor_compare}

\section{Experiment}

\input{table/model_deploy_eff}

\subsection{Clinical Deployment Efficiency}
As shown in Table \ref{tab:model_deploy_performance}, FundusExpert-mini(1B), optimized for consumer GPUs, excels in accuracy and efficiency over models like InternerVL2.5-38B. On an RTX 4090, it achieves 0.20 img/s (2.0GB VRAM, bs=1), scaling to 2.34 img/s (max BS 128) (Table~\ref{tab:model_deploy_performance}). In contrast, larger models like InternerVL2.5-38B are less accurate, require high-end A100 GPUs, support very limited batch sizes (Max BS 2), and have slow inference. FundusExpert-mini provides an optimal balance for widespread clinical adoption.

\subsection{Performance Evaluation}

\subsubsection{Extrapolation Ability of FundusExpert} 

\begin{figure*}[ht]
\centering
\includegraphics[width=\textwidth]{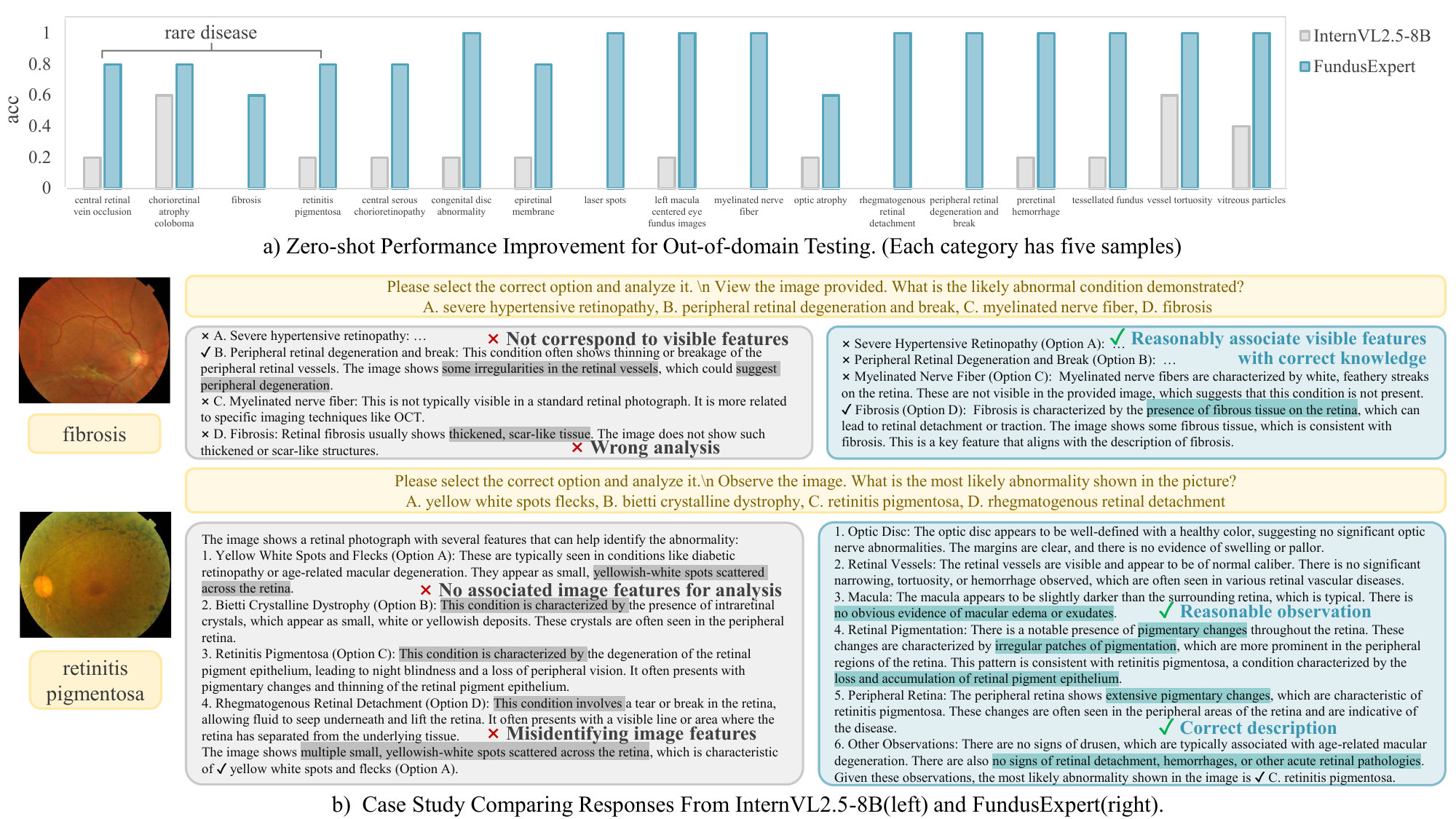}
% \vspace{-0.2cm}
\caption{Comparison of the foundation model and FundusExpert on out-of-distribution representative categories. }
\label{fig:GMAI_bar_2_example}
% \vspace{-5mm}
\end{figure*}
 
FundusExpert demonstrates extrapolation reasoning ability in out-of-domain tasks on GMAI-MMBench. As shown in Table 1, it achieves a 66.7\% accuracy rate in zero-shot tasks on GMAI-MMBench, surpassing the base model InternVL2.5 by 30.2\%. This is primarily attributed to FundusGen's explicit modeling of clinical feature inference logic. Case comparisons in Figure \ref{fig:GMAI_bar_2_example} further validate this ability. 

For the "retinitis pigmentosa" diagnostic task in Figure \ref{fig:GMAI_bar_2_example}(b), FundusExpert locks onto the correct diagnosis based on the peripheral retinal pigment deposition pattern through extensive feature analysis and exclusion of other options, while the pretrain model incorrectly identifies the features.

\subsubsection{Evaluation of Zero-shot Ability in Open-domain Tasks}

\noindent\textbf{Localization Ability Evaluation.} 

Under evaluation, the IoU calculation formula is:

\begin{equation}
    IoU = \frac{\text{TP}}{\text{TP} + \text{FP} + \text{FN}},
\end{equation}
where \(\text{TP}\) is the number of intersection pixels between the predicted box and the ground truth region, \(\text{FP}\) is the number of redundant pixels in the predicted box that exceed the ground truth region, and \(\text{FN}\) is the number of missed pixels in the ground truth region that are not covered by the predicted box.

\noindent\textbf{Clinical Consistency Evaluation in Medical Report Generation.} 

\begin{figure}[ht]
\centering
\includegraphics[width=0.47\textwidth]{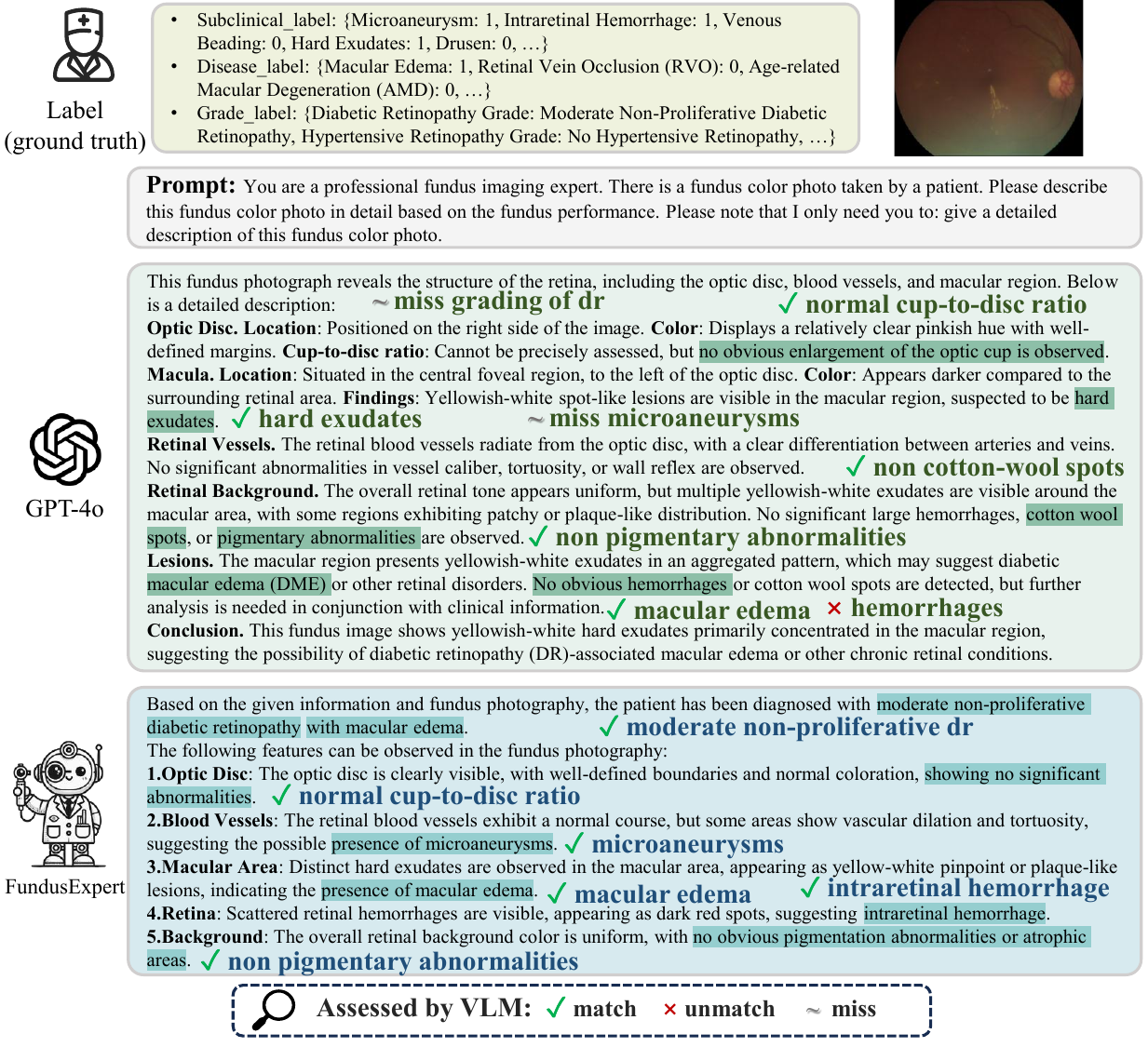}
\caption{Example of Clinical Consistency Evaluation}
\label{fig:report_gen}
\end{figure}

We propose a multi-granularity semantic matching framework to compute the accuracy of medical report generation tasks. It uses VLM(GPT-4o) to decouple the structured evaluation of clinical logical consistency in generated reports.

Existing likelihood-based benchmarks for medical text generation, such as BLEU and ROUGE, inadequately assess semantic plausibility. To overcome this, we introduce a multi-granularity semantic matching framework that evaluates the accuracy of generated medical reports. This framework leverages a Vision Language Model (VLM), specifically GPT-4o, to perform a structured evaluation of clinical logical consistency.

Let the set of ground-truth labels be $\mathcal{L} = \{l_1, l_2, ..., l_N\}$, which includes both positive and negative findings. Let the set of semantic features extracted from the generated report be $\mathcal{S} = \{s_1, s_2, ..., s_M\}$. The clinical consistency score is defined as:

\[
\text{Clinical Consistency} = \frac{\sum_{i=1}^N \mathbb{I}(\text{match}(l_i, \mathcal{S}))}{|\mathcal{L} \cup \mathcal{S}|}
\]
where, the function $\text{match}(l_i, \mathcal{S})$ checks for a bidirectional semantic correspondence between a label $l_i$ and the set of generated features $\mathcal{S}$, as determined by the VLM. $\mathbb{I}(\cdot)$ is the indicator function, which is 1 if the condition is true and 0 otherwise. The denominator $|\mathcal{L} \cup \mathcal{S}|$ is the size of the union of the ground-truth labels and the generated features(determined by the VLM), which normalizes the score.

% Where:
% \begin{itemize}
%     \item The function $\text{match}(l_i, \mathcal{S})$ checks for a bidirectional semantic correspondence between a label $l_i$ and the set of generated features $\mathcal{S}$, as determined by the VLM.
    
%     For a positive label $l_i$, a match occurs if the report's semantics $\mathcal{S}$ correctly describe the finding. And for a negative label $l_i$, a match occurs if the report's semantics $\mathcal{S}$ correctly state the absence of the finding.
%     \item $\mathbb{I}(\cdot)$ is the indicator function, which is 1 if the condition is true and 0 otherwise.
%     \item The denominator $|\mathcal{L} \cup \mathcal{S}|$ is the size of the union of the ground-truth labels and the generated features(determined by the VLM), which normalizes the score.
% \end{itemize}

FundusExpert achieves 77.0\% in clinical consistency evaluation, significantly outperforming GPT-4o, which scores 47.6\% (+29.4\%). This advantage stems from the model's ability to model multi-level pathological associations. For example, in diabetic retinopathy report generation, the model not only accurately identifies microaneurysms and macular edema but also verifies the stage of the lesion through contextual semantics, such as the distribution of retinal hemorrhages.

\subsection{Supplement to the ablation experiment results}

\noindent\textbf{Cognitive Chain Construction Data Ablation.}
The results of the clinical QA task evaluation are shown in Table 3. The average diagnostic accuracy for diseases in the GMAI-MMBench decreases by 3.5\% for the (2) Cognitive Chain Degradation group compared to (1). 

Further analysis reveals that the average diagnostic accuracy for 21 complex diseases, such as retinitis pigmentosa, in the GMAI-MMBench decreases by 4.8\% (75.2\% $\to$ 70.4\%) for the (2) Cognitive Chain Degradation group compared to (1), indicating that reasoning by constructing a progressive chain, enhances the model's logical deduction ability for complex pathologies.
These 21 diseases include 17 of the rarer disease categories in Figure \ref{fig:GMAI_bar_2_example} as well as bietti crystalline dystrophy, fundus neoplasm, vkh disease, and pathological myopia. The prevalence of these diseases is relatively low, or they represent more severe or specific pathological conditions than common diseases (such as common myopia, cataracts).

\noindent\textbf{Startup Data Ablation.}
Startup data enhances the model's basic understanding of different diseases by providing diverse disease descriptions.
In addition to the performance degradation in Table 3 (training for 1 epoch), further experiments show that (6) requires 0.5 additional epochs (training for 1.5 epochs) to achieve the same accuracy as (5) on Fundus-MMBench, indicating that there is a delay in convergence without startup data. At the same time, its performance on out-of-distribution GMAI-MMBench worsens, with the gap increasing from $\downarrow$4.5\% to $\downarrow$5.9\%.

%% file: table/preudo-label-num.tex
\begin{table*}[ht]
    \centering
    \small
    \setlength{\tabcolsep}{8 pt} 
    \renewcommand{\arraystretch}{1.2}
    \begin{tabular}{lccccc}
        \toprule[0.4mm]
        \textbf{Dataset} & \textbf{Microaneurysms} & \textbf{Hard Exudates} & \textbf{Cotton-wool Spots} & \textbf{Optic Cup} & \textbf{Optic Disc} \\
        \midrule[0.3mm]
        Open-source Dataset (True Labels) & 882  & 642  & 291  & 901  & 1070 \\
        Annotated Dataset (Pseudo-labels)        & 5357 & 10089 & 1876 & 16551 & 16720 \\
        \bottomrule[0.4mm]
    \end{tabular}
    \caption{Comparison of label quantities between open-source and annotated datasets.}
    \label{tab:Pseudo_nums}
\end{table*}

%% file: table/model_deploy_eff.tex
\begin{table*}[htbp]
  \centering
  % \small
  \caption{Model deployment efficiency comparison. Metrics include Accuracy (Acc.) on Fundus-MMBench and GMAI-MMBench, Throughput (Thrpt.) in images per second (img/s), VRAM Memory (Mem.) in GB at batch size 1 (bs=1), Maximum deployable Batch Size (Max BS), and Throughput at Max BS. Results highlighted in gray were obtained on an RTX 4090; all other results were obtained on an A100 GPU.}

  \label{tab:model_deploy_performance}
  \setlength{\tabcolsep}{11.5 pt} 
  \renewcommand{\arraystretch}{1.0}

 \begin{tabular}{lc cccccc}
    \toprule[0.4mm]
    Model & Params & Fundus & GMAI & Thrpt. & Mem.(bs1) & Max & Thrpt.(Max) \\
    &   Num    & Acc(\%)& Acc(\%)& (img/s) & (GB) &  BS & (img/s) \\
    \midrule
    InternerVL2.5-8B    & 8B   & 30.6 & 37.8 & 0.14 & 16.3 & 32  & 1.43 \\
    InternerVL2.5-38B   & 38B  & 44.0 & 42.3 & 0.03 & 74.0 & 2   & 0.07 \\
    \midrule
    FundusExpert-mini   & \textbf{1B}   & 63.5 & 58.3 & \textbf{0.24} & \textbf{2.0}  & \textbf{512} & \textbf{3.54} \\
    FundusExpert        & 8B   & \textbf{69.7} & \textbf{66.7} & 0.14 & 16.3 & 32  & 1.43 \\
    \midrule
    % \cellcolor{gray!20} FundusExpert-mini & \cellcolor{gray!20} \textbf{1B}   & \cellcolor{gray!20} 63.5 & \cellcolor{gray!20} 58.3 & \cellcolor{gray!20} \textbf{0.20} & \cellcolor{gray!20} \textbf{2.0}  & \cellcolor{gray!20} \textbf{128} & \cellcolor{gray!20} \textbf{2.34} \\
    % \cellcolor{gray!20} FundusExpert      & \cellcolor{gray!20} 8B   & \cellcolor{gray!20} \textbf{69.7} & \cellcolor{gray!20} \textbf{66.7} & \cellcolor{gray!20} 0.10 & \cellcolor{gray!20} 16.3 & \cellcolor{gray!20} 4   & \cellcolor{gray!20} 0.31 \\
    
    \rowcolor{gray!20} % Light gray background for RTX 4090 results
    FundusExpert-mini & \textbf{1B}   & 63.5 & 58.3 & \textbf{0.20} & \textbf{2.0}  & \textbf{128} & \textbf{2.34} \\
    \rowcolor{gray!20} % Light gray background for RTX 4090 results
    FundusExpert  & 8B   & \textbf{69.7} & \textbf{66.7} & 0.10 & 16.3 & 4   & 0.31 \\
    \bottomrule[0.4mm]
  \end{tabular}
\end{table*}